\documentclass[10pt]{article}
\usepackage{graphicx,psfrag,amsmath,amsfonts,verbatim}

\usepackage{arxiv}

\usepackage[utf8]{inputenc} 
\usepackage[T1]{fontenc}    

\usepackage{hyperref} 
\usepackage{url}            

\usepackage[font={small,it}]{caption}

\usepackage{booktabs}       
\usepackage{amsfonts}       
\usepackage{nicefrac}       
\usepackage{microtype}      
\usepackage{lipsum}
\usepackage{cite}
\usepackage{calc}

\usepackage{amsmath}
\usepackage{setspace}
\usepackage{enumitem}
\usepackage{mathpazo}

\usepackage{longtable}
\usepackage{xcolor}
\usepackage{soul}
\usepackage{nomencl}
\usepackage{float}
\makenomenclature

\title{\Large{System-Level Development of a User-Integrated Semi-Autonomous Lawn Mowing System: Problem Overview, Basic Requirements, and Proposed Architecture}}

\author{
  Albert E. Patterson, Yang Yuan, and William R. Norris\\
  \textit{Department of Industrial and Enterprise Systems Engineering}\\
  University of Illinois at Urbana-Champaign, 
  Urbana, Illinois 61801 \\
  Correspondence: \texttt{\url{wrnorris@illinois.edu}} \\
}

\begin{document}
\maketitle

\begin{abstract}
This concept paper outlines some recent efforts toward the design and development of user-integrated semi-autonomous home-sized lawn mowing systems from a systems engineering perspective. This is an important and emerging field of study within the robotics and systems engineering communities. The work presented includes a review of current progress on this problem, a discussion of the problem from a systems engineering perspective, a general system architecture developed by the authors, and a preliminary set of design requirements. This work is meant to provide a baseline and motivation for the further development and refinement of these systems within the systems engineering and robotics communities and is relevant to both academic and commercial research.
\end{abstract}


\section{Introduction}
\label{sec:Sec1}
\vspace{-5pt}
Lawn mowing is one of the most essential tasks required for most home owners, requiring a significant investment of time and resources to accomplish properly. In 2018, lawn mowing equipment was estimated to be a US \$27.5 billion global industry and projected to continue growing a further 5.2\% between 2019 and 2025~\cite{GVR2019}. In addition to the significant investment in equipment and time for the users, it is also typically considered to be an unpleasant task to complete and one that users would like to avoid when possible. In recent years, this has given rise to attempts to make lawn mowers more robotic and more autonomous, resembling the now-widespread Roomba vacuum cleaners~\cite{Tribelhorn2007,AponteRoa2019,HicksII2000}. There are a number of technical and system-level challenges with doing this for lawn mowers, however; this is a much more complex problem than the development of a robotic vacuum cleaner for four major reasons, namely, (1) the boundary and environmental conditions during use are much less well-defined, (2) a lawn mower has the potential to cause significant damage and injury to humans and animals if is goes rogue, (3) it is a much larger system with more complex set of sensors and controls, and (4) it is a much more valuable and expensive system. In addition to driving complexity in the system, these reasons all necessitate a rigorous security system, including hardware and software security, and a remote kill switch. To support the mower system, an effective method for maintaining the system is needed, as well as a mapping method.    

\begin{figure}[H]
\centering
\includegraphics[width=0.8\textwidth]{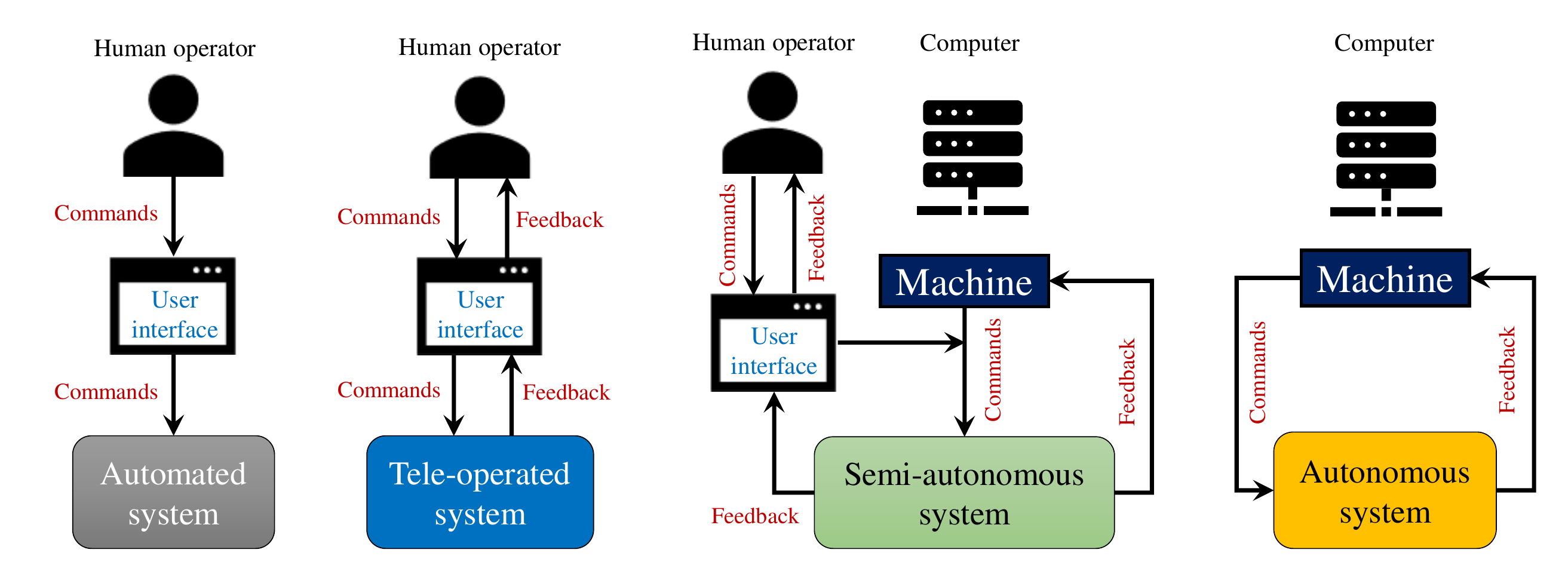}
\caption{Automated/tele-operated, semi-autonomous, and autonomous system concepts}
\label{fig:Aut}
\end{figure}

The application of autonomous and semi-autonomous robotic systems to lawn mowing and other human-assistant tasks~\cite{Sahin2007, Gregg2006} has been the topic of several major studies, mainly focused on how to accomplish effective path planning and how to control the system. Autonomous and semi-autonomous mowers should be clearly distinguished from simple automatic/robotic mowers~\cite{HicksII2000} (such as tele-operated systems or Roomba-like systems), as the latter are not the topic of the present study. As it is typically understood, an autonomous system is defined as one that has no human interaction or decision-making in response to unexpected events encountered by the system during operation; in the case of a semi-autonomous system, there is a degree of autonomy in the system, but there is still a human interface with the system and the operator provides at least some of the control and direction of the system. At the current time, full autonomy that is effective, reliable, and ethical remains an ideal that is yet to be accomplished in practice without some kind of human monitoring during operation. A logical understanding of the theoretical relationship between autonomous, semi-autonomous, and automated/tele-operated systems is shown in Figure~\ref{fig:Aut}. These definitions are given in order to ensure that the approach and architecture presents in this paper can be clearly understood and contextualized. 

Toward the development of autonomous/semi-autonomous mowing systems, Shiu \& Lin~\cite{BingMinShiu2008} considered several different modes for optimal path planning, including minimal working time, minimal energy consumption, and mixed operation. The system required the use of GPS for localization and tracking of the job. Considerations for path planning and obstacle avoidance in area where GPS coverage is poor was explored in depth by Nourani-Vatani et al.~\cite{Nourani-Vatani2001}; localization was done using laser-based localization and a virtual force field algorithm near identified obstacles. There was some significant deviation noted from the planned paths during physical experiments, but the method did effectively avoid the obstacles. Focus only on the development of an effective autonomous/semi-autonomous mower controller was the topic of studies by Wasif et al.~\cite{Wasif2011} and Daltorio et al.~\cite{Daltorio2010}. The controller developed by Wasif et al. used sonar to find the ranges to any identified obstacles, GPS to localize the mower in the global space, local positioning using odometry, and an on-board camera to track the mowed areas. Daltorio used a similar range of instruments, with the addition of LIDAR and a greater focus on GPS positioning. 

Work by Peless et al.~\cite{Peless2001} developed a local navigation system that utilized a proximity sensor on the robot itself which was able to interact with a defined boundary on the working area; the robot would begin at a known location and then continuously track its location relative to the boundary. Corrections during used were made by comparing the measured coordinated with those on a user-defined map. Generation of the map automatically using UAVs was explored by Samad et al.~\cite{Samad2013}, while the use of ground robots for this purpose was proposed by Choi et al~\cite{Choi2012}. In both studies, high-quality digital images were used to generate a map of terrain and potential obstacles, while the ground robot was supplemented with LIDAR to aid in the detection of important terrain features and partially hidden obstacles. These methods were found to be effective in locating grassy areas and obstacles but still missed some features such as grates and manhole covers. A practical configuration for mapping and avoiding obstacles was proposed by Franzius et al.~\cite{Franzius2017}, which used a single camera and simple processing board which could be added to existing robotic mowing systems. It was observed over extensive testing (nearly 3,500 hours of testing) that the system developed did not need any direct human intervention to accomplish acceptable mowing performance, but some collisions with small obstacles did occur as the system was not able to locate and avoid all of them.  

Processing of the images collected by the mapping strategies is another topic of research, as effective machine interpretation of the images and conversion into a map is vital to the working of an autonomous/semi-autonomous mowing system. Schepelmann et al~\cite{Schepelmann2010} proposed an image-segmentation method which extracts identifiers to locate grass-containing regions and then apply use the texture statistics for distinguishing grass and obstacles. This method is inexpensive and quick, but may be biased by unfavorable lighting conditions or blurry areas in the used images. This texture division method was further developed by Guo et al.~\cite{Guo2017}, making the lighting and scaling factors of the images much less impactful through the use of a Gabor filter. Zhang et al.~\cite{Zhang2018} proposed a method for estimating the biomass of grass within a specific area from images, helping to estimate its weight and density; this was done primarily for predicting fire control needs, but the method is applicable to mowing, as it can help calculate mowing time, power requirements, and other important characteristics during job planning.      

Automated and intelligent path planning, as long as it can be shown to be reliable for the needed conditions, can help to find the best coverage for a particular yard or field. These methods may be used to adapt in real-time if the map needs to be updated after beginning the job (such as when the system encounters an unexpected obstacle). While a reliable and efficient method for accomplishing this does not seem to have been developed yet, there are several established solutions that work well in particular applications. The most popular approach to the problem uses a divide-and-conquer strategy~\cite{Choset2001} to achieve completeness easily by dividing a given area into several cells and covering each cell by a path that requires minimum time. Choset~\cite{Choset1998,Choset2001} introduced boustrophedon decomposition, where a line segment is swept through a given environment, forming a new cell when there is a change in the connectivity of the slice described by critical points. The robot then covers each of the cells using a simple zig-zag motion. Acar et al.~\cite{Acar2002} used Morse decomposition to use the critical points of Morse function to form the cells so that complete coverage of each cell is guaranteed. A cell decomposition algorithm based on topological structures was proposed by Wong \& MacDonald~\cite{Wong2003}, which is more simple and time-efficient than many other methods. To save time during operation and reduce the distance traveling by the robot, Yao~\cite{Yao2006} tried to ensure that the entry and exit points between each of the cells coincided or were at least as close as possible to each other. 

While much previous work has been completed in the areas of path planning, navigation and obstacle avoidance, as well as development of complete semi-autonomous systems, there is need of a high-level, systems-focused approach to the development of these systems. A common theme in all of the reviewed studies was that they were approaching the problem as a sort of "new" area of study, resulting in a plethora of different approaches and wide variance in the level of success in different environments. It should be noted that a large number of the previous papers in this area are published in conference proceedings and not archival journals, which may make the status of the problem more difficult to understand since the proceedings are less organized and less available (in general) than journal articles would be. It could dramatically increase the pace of development if there was some kind of standard or development framework for these systems. Systems engineering theory could provide a good approach for accomplishing this, as it can provide a general "fill-in-the-blank" framework that is easy to organize and understand. The development of a unified approach to the design and analysis of mowers and other semi-autonomous systems will also greatly aid in the testing and certification of these systems.   

\begin{figure}[H]
\centering
\includegraphics[width=0.5\textwidth]{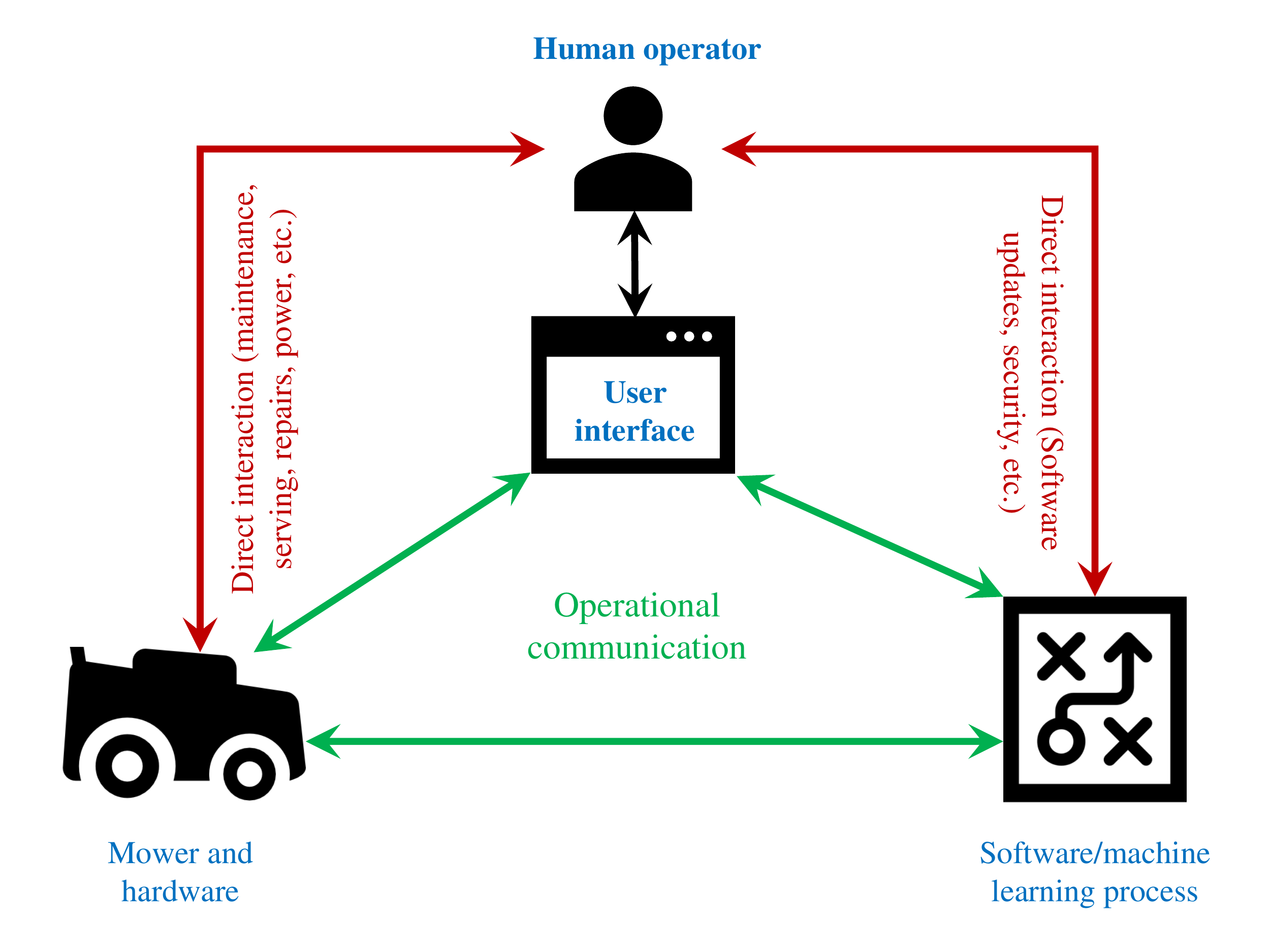}
\caption{Semi-autonomous mowing system top-level interfaces}
\label{fig:Interfaces}
\end{figure}

One of the most important aspects of the system engineering approach will involve understanding the interfaces in the semi-autonomous mowing system, most especially the interface between the human user and the rest of the system. While there will be a number of subsystems within the whole framework, the three main interfaces at the top level will be between the human user, the system hardware, and the software (Figure~\ref{fig:Interfaces}). The rapid development in recent years of mobile technology provides a great opportunity for the further development of efficient and realistic consumer-level semi-autonomous lawn mowing systems. Advanced mobile platforms provide a very good natural interface between the human user and can be used to increase the confidence of users in the system. While the semi-autonomous mowing system may be a new and unfamiliar technology for the user, the mobile-based user interface could help the user learn how to use the system and making them feel more safe and familiar with it.      

The purpose of this concept paper is to provide a high-level, systems engineering perspective on the design and development of a general user-integrated semi-autonomous mowing system. The concept of operations (CONOPS) and the general systems engineering approach will be discussed in Section~\ref{Sec2}, followed by an examination of the basic system requirements in Section~\ref{Sec 3}. In Section~\ref{Sec 4}, a high-level, generic architecture for such a system is proposed and examined from several perspectives. Finally, conclusions and ideas for future work will finally be given in Section~\ref{Sec 5}. This work is meant to be a concept paper to help organize the work and provide a baseline for further work in this area. This is not a complete research paper nor does it claim to provide a full design and certification process for a specific user-integrated, semi-autonomous mowing system.     

\section{Concept of Operations (CONOPS)}
\label{Sec2}
\begin{figure}[H]
\centering
\includegraphics[width=0.8\textwidth]{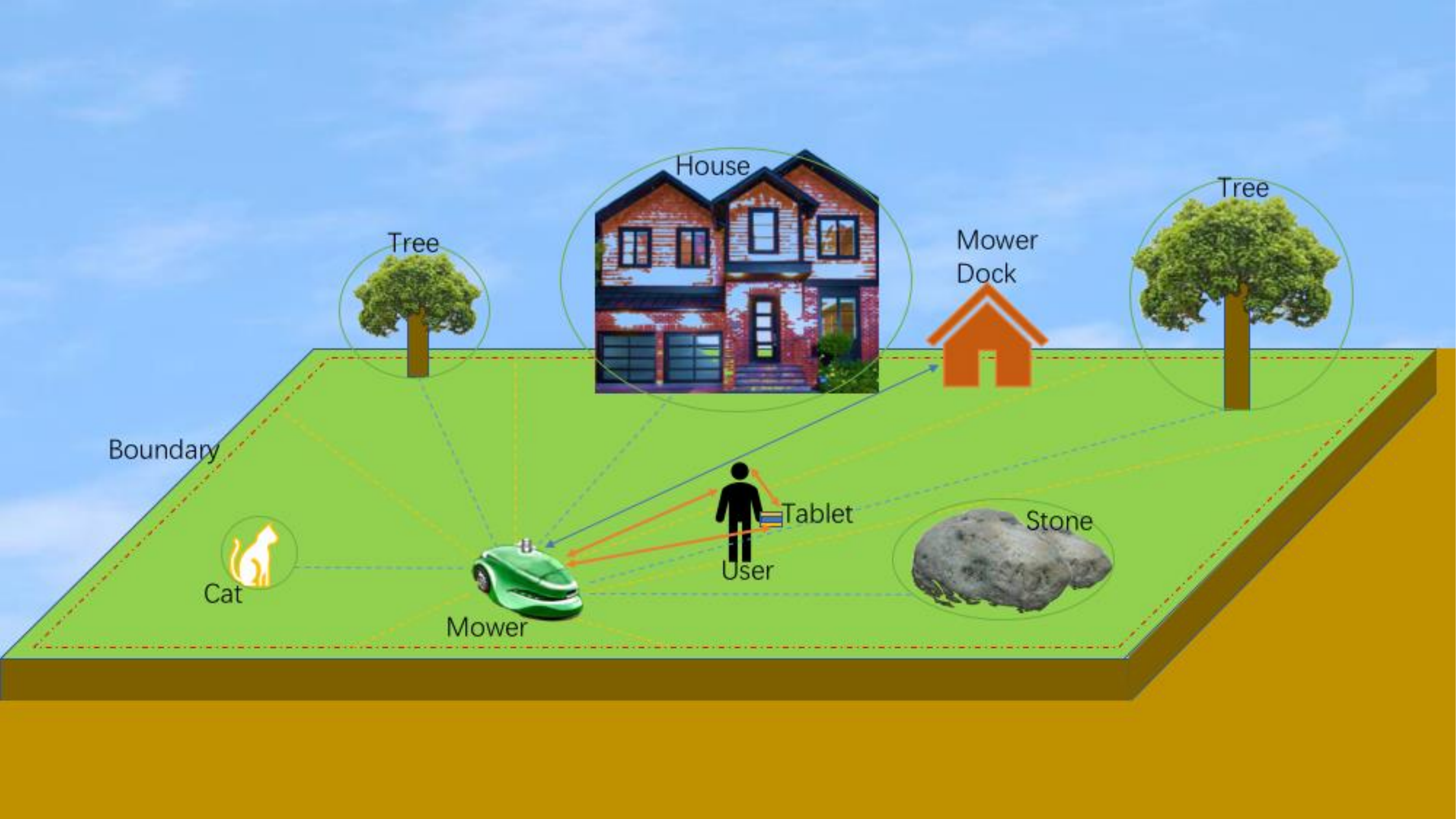}
\caption{Concept of operations (CONOPS) for home-based semi-autonomous lawn mowing system~\cite{Norris2019}}
\label{fig:CONOPS}
\end{figure}

\vspace{-5pt}
The basic CONOPS for a general user-integrated semi-autonomous mowing system is shown in Figure~\ref{fig:CONOPS}. It is assumed that this kind of system is primarily for residential use and that a yard with a house and predicable and unpredictable obstacles is the operational space for the system. The basic elements of this operational space are the user and user interface (represented by a tablet), the mower, the ground to be mowed (i.e, the yard map), the base of operations (represented by the house), the mower base (charger, fuel, maintenance, etc., represented by the mower dock), the yard boundary, and a set of obstacles. The obstacles can be divided into three categories: (1) the predictable non-interfering obstacles (PNIOs) (represented by the rock), (2) the predictable interfering obstacles (PIOs) (represented by the trees), and (3) the unpredictable obstacles (represented by the cat). 

The mowing system must have some method of localizing itself so that it can track its position and progress toward completion of the mowing job. The yard map may be generated by the user, by a UAV or ground tracking robot, or some other method; depending on the local conditions and the geometry of the yard, it may be necessary to map the yard at the beginning of each mowing job or it may only need to be done once when the mowing system is started for the first time. The map should be converted into an occupancy grid (statistical or binary) or other machine-readable data that the system can use to path plan and avoid obstacles. It is assumed for this kind of system that an on-board computer will be used on the mower to process all the needed data, run the autonomous part of the system, communicate with the user, and run the basic mower functions.

It must also be able to react to obstacles in some way, whether predicted obstacles (i.e. marked on the yard map) or unpredicted obstacles. The unpredicted obstacles (such as small animals, children's toys, or other things will be missing from a standard mapping of the yard) must be tracked using some means, whether that is the use of a camera or LIDAR or whether the user is directly involved if such as obstacle is encountered. The predicted obstacles are already known to the system before mowing commences (via the occupancy grid or other map), as they were located by the mapping system or marked by the user. These will consist of PNIOs, which are things such as rocks, ponds, yard ornaments, and other things that are not tall enough or dense enough to interfere with the navigation and localization of the system. On the other hand, PNOs, such as cars, trees, and buildings, will likely interfere with the operation and communication of the system by blocking signals or GPS coverage. Care should be taken when setting up the system (e.g., placing the dock or locating beacons and kill wires for the system, and similar) to avoid these kinds of obstacles by default when possible. When using a UAV-based mapping, a feature of the analysis should be to locate any obstacles or potential obstacles that could interfere with signalling and localization and classify them as such.

\section{Basic, High-Level Requirements Development}
\label{Sec 3}
\begin{figure}[H]
\centering
\includegraphics[width=1\textwidth]{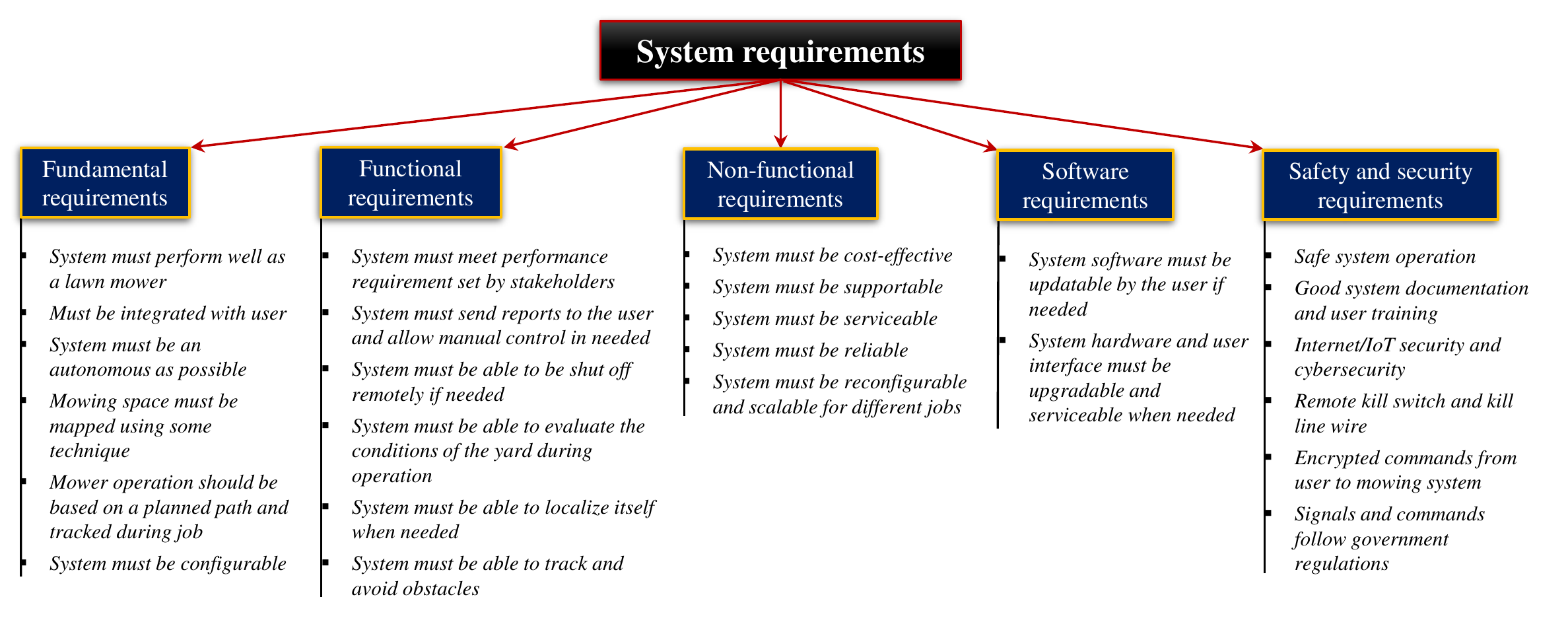}
\caption{Basic top-level requirement categories for a general user-integrated semi-autonomous mowing system with example lower-level requirements}
\label{fig:Requirements}
\end{figure}

\vspace{-5pt}
The design and setup of any user-integrated semi-autonomous mowing system should follow a set of fundamental requirements, all of which are based on the needs and safety of the system and are not driven by a specific system. These standardized basic requirements are the starting place for the development of a system-specific set of requirements, driven from the needs and desires of the stakeholders. These top-level starting requirements can be divided up into five categories, namely (1) the fundamental requirements, (2) functional requirements, (3) non-functional requirements, (4) software requirements, and (5) the safety and security requirements. Figure~\ref{fig:Requirements} shows the tree diagram for these requirements with some example lower-level requirements to illustrate the relationship; note that these are for example only and the true system requirements will depend on the stakeholders needs and the operational conditions of the system being designed. The purpose of this outline is to give a starting place which will guide the development of the true system requirements in an organized, consistent, and (eventually) standardized way.

As shown in Figure~\ref{fig:Requirements}, the best place to begin the development of a requirements document for this kind of system is to identify or specify the main requirements in each of the essential areas. 
\begin{itemize}
    \item \textbf{Fundamental requirements}: These are the most important high-level requirements, such as the expected basic performance and configuration of the system. In the case of a user-integrated semi-autonomous mowing system, these requirements would be things such as requiring a basic user interface, that the system should have a significant degree of autonomy, that the system would function using a planned path and avoid obstacles, and that the space to be mowed must be mapped somehow and have a boundary identifiable by the mowing system. 
    \item \textbf{Functional requirements}: Since the fundamental requirements (i.e., "the system must do X"), the functional requirements are those related to performance and function of the system. These may include things such as meeting minimal performance and efficiency requirements, being able to be remotely shut off in an emergency, being able to localize itself once the job is started, and other functional requirements. 
    \item \textbf{Non-functional requirements}: These would contain the bulk of the general requirements and deal with things such as reliability, serviceability, supportability, and cost effectiveness. This list of requirements could be the most broad, as these may relate to things external to the system, such as local regulations and the desires of the user not related to mowing quality (e.g., color, noise level, IOS versus android user platform, etc.). 
    \item \textbf{Software requirements}: These requirements would be separate from the other requirements, as the software used will likely be developed separate from the rest of the system and may be used in a wide variety of systems, similar to 3-D printer software packages such as Cura\textsuperscript{\textregistered} and Repetier-Host\textsuperscript{\textregistered}. The software development may also have a large influence on the development of the rest of the system, as it may impose operational constraints on the rest of the system. However, whether the software is open-source or proprietary, it should allow upgrades and bug and security fixes either by the user or pushed by the manufacturer. 
    \item \textbf{Safety and security requirements}: It is vitally important for the system to be safe to use for the operator, any by-standers, and any animals or property that could be encountered by the mower. In addition, the system will need to connected to the internet (or at least a local network) and will have high-value hardware components, it is vital that robust physical and cyber security is implemented. 
\end{itemize}
A more extensive and detailed suggested list of requirements for this kind of system can be found in the technical report "User-Integrated Semi-Autonomous Lawn Mowing Systems:
Example Basic, Functional, Non-Functional, and Safety and Security Requirements" by the authors of this concept paper~\cite{Norris2019}. The technical report was meant to supplement this work and should be used in connection with it.

\section{Proposed System Architecture}
\label{Sec 4}
\vspace{-5pt}
The physical architecture of a user-integrated, semi-autonomous mowing system will depend heavily on the specific design requirements developed by the stakeholders. However, the basic top-level architecture can be standardized if some fundamental system characteristics are used as the starting point in any design. These may be organized in a variety of ways, but the relationship between the essential items will remain the same regardless of the final detained design. In particular, 
\begin{itemize}
    \item The user will interact with the working system through the user interface
    \item The system will operate with some degree of autonomy with the user providing some of the monitoring, direction, and decision making
    \item The hardware and software elements must interact through the sensors, the on-board computer, and the localization system
    \item There must be some boundary elements for the yard for the localization engine and sensors to interact with
    \item There must be some kind of secondary vision system in order to detect and react to unpredictable obstacles
    \item There must be some method of producing a map of the yard, which will include locating and identifying PNIOs and PNOs
    \item There must be a communication link between the system and the internet or a local wireless network
    \item The system must have some kind of home base or dock for it to be secured, maintained, updated, recharged or refueled, and similar
\end{itemize}

\begin{figure}[H]
\centering
\includegraphics[width=0.9\textwidth]{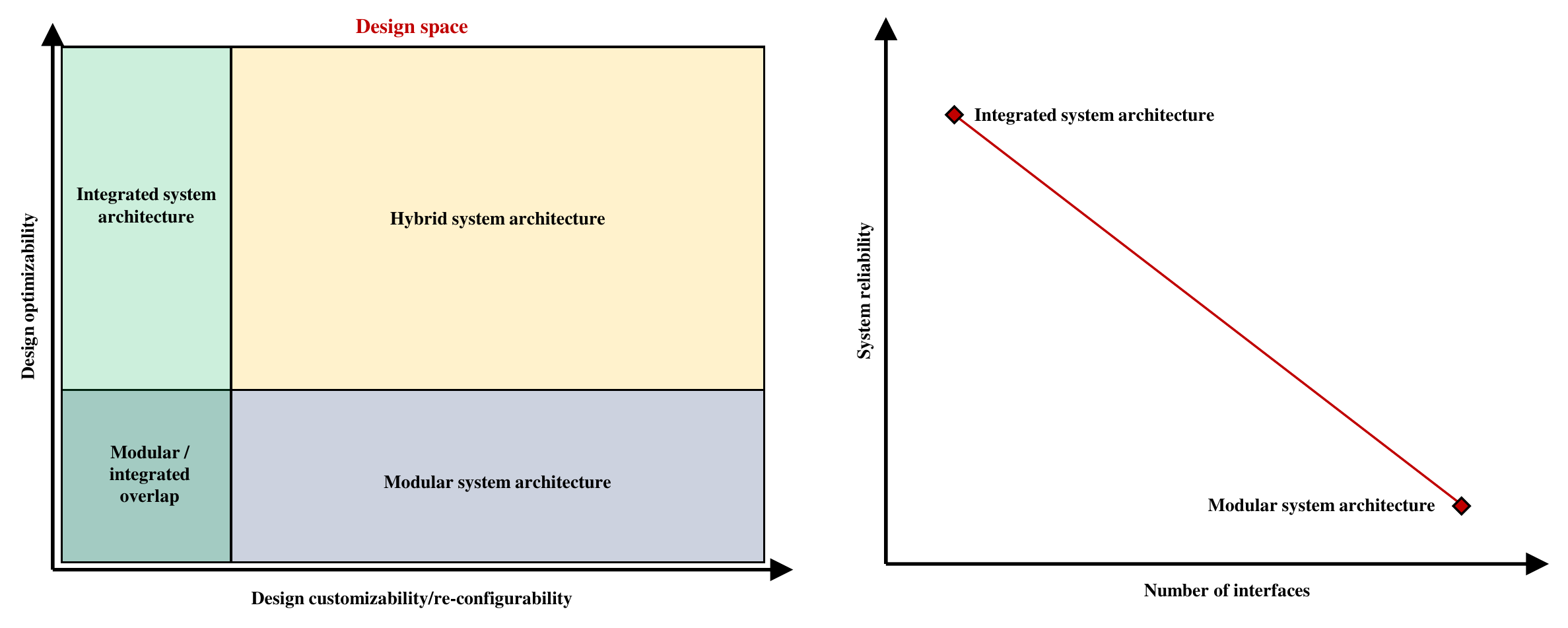}
\caption{Modular, integrated, and hybrid architecture complexity and reliability}
\label{fig:ModInt}
\end{figure}

These characteristics are important for any system design, so a logical way to understand and organize them from a system perspective will be helpful in standardizing the system design process. Therefore, this paper proposes a high-level system architecture which divides the system unto a series of parts and interfaces. Three cases are possible, namely (1) a modular system, (2) an integrated system, and (3) a hybrid system. The basic relationship between the three configurations related to the optimizability and system customizability/re-configurability is shown in Figure~\ref{fig:ModInt}. Note that (also shown in Figure~\ref{fig:ModInt}) the system reliability will be generally inversely correlated with the number of interfaces in the system, as it is well-known in the systems engineering literature that the system interfaces are the typical points of failure in most systems~\cite{Murthy2009,Aal2012,Elsayed2012,Hohenbichler1982,Chern1992,Frey2007,ElMaraghy2012,Xu1990,Billinton1983}. This is especially true with a modular system like the one described in this paper, since most of the components/subsystems will be connected in a series configuration. This is of course a very simplified view of the system and the actual degree of influence will depend on many factors in the final system design itself. It is however a very important consideration that cannot be ignored when selecting what basic system architecture to use. 

Clearly, there is a trade-off for using either basic architecture and which is appropriate is dependent on the needs and wished of the stakeholders; however, it is the opinion of the authors that a hybrid system will usually be the most appropriate choice. Note that in any case, the system hardware and software will be considered different subsystems, as realistically they will likely be developed by separate design teams and later integrated together. In addition, it would be very difficult to integrate the user interface and network and support systems into the main system. Therefore, in all three cases, there will be four divisions of the system at the top architecture level.  

In a pure modular system, the architecture is divided into a series of subsystems, which are in turn composed of modular hardware and software components down to the level of the individual element. However, this type of system is rare and usually very inefficient since all components must be modular, easily replaceable, and easily available for replacement. This prevents much optimization of the system, but gives the highest level of system flexibility. Designing such a system with strictly standardized parts and open-source hardware and software may provide an attractive option to some customers who would use it as a hobby project. However, with the current status of the autonomous vehicle technology and the safety hazards involved, this would likely present too much liability for a consumer product.

\begin{figure}[H]
\centering
\includegraphics[width=1\textwidth]{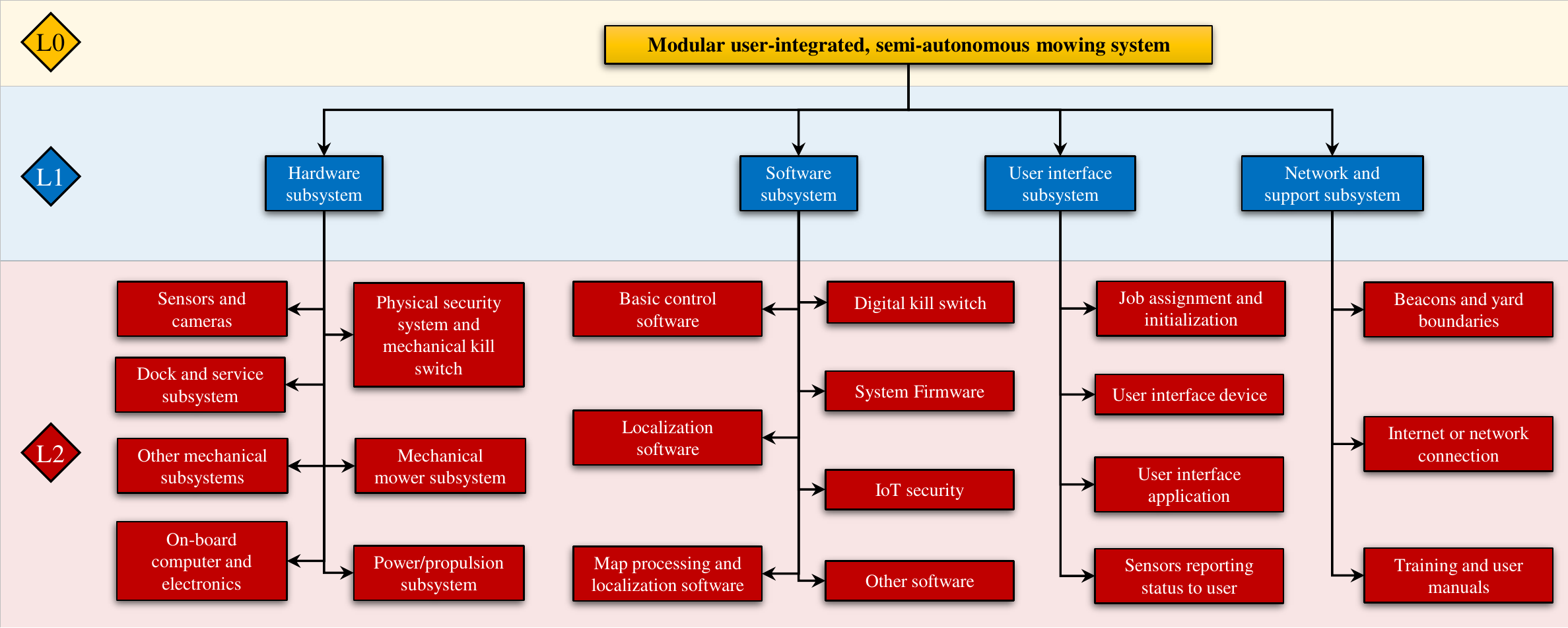}
\caption{Proposed modular system architecture (first three levels)}
\label{fig:Modular}
\end{figure}

In this case, the top four subsystems would consist of the hardware subsystem, the software subsystem, the user interface subsystem, and the network and support subsystem; Figure~\ref{fig:Modular} shows the first three levels of the architecture, enough to establish the system configuration and basic layout. Note that each of the red (L2) items may be individual components or subsystems depending on the final design and the final set of design and performance requirements set by the stakeholders. Within the hardware subsystem would be the mechanical mower components, the power/propulsion source, the sensors, the on-board computer, the dock, and similar mechanical items. The software subsystem would contain the base control software (itself perhaps a modular build), the localization software, the mapping and map processing software, the digital kill switch, the firmware for the system, and other important software parts and subsystems. The user interface subsystem will consist of things related to the user experience and control of the system, containing both hardware and software components such as the interface device, the application software to run it, and the sensors to give updates and reports to the user as the job is being completed. Finally, the network and support subsystem consists of the user training materials, the beacons/navigation markers for the system, and the network connection for the mower.     

\begin{figure}[H]
\centering
\includegraphics[width=1\textwidth]{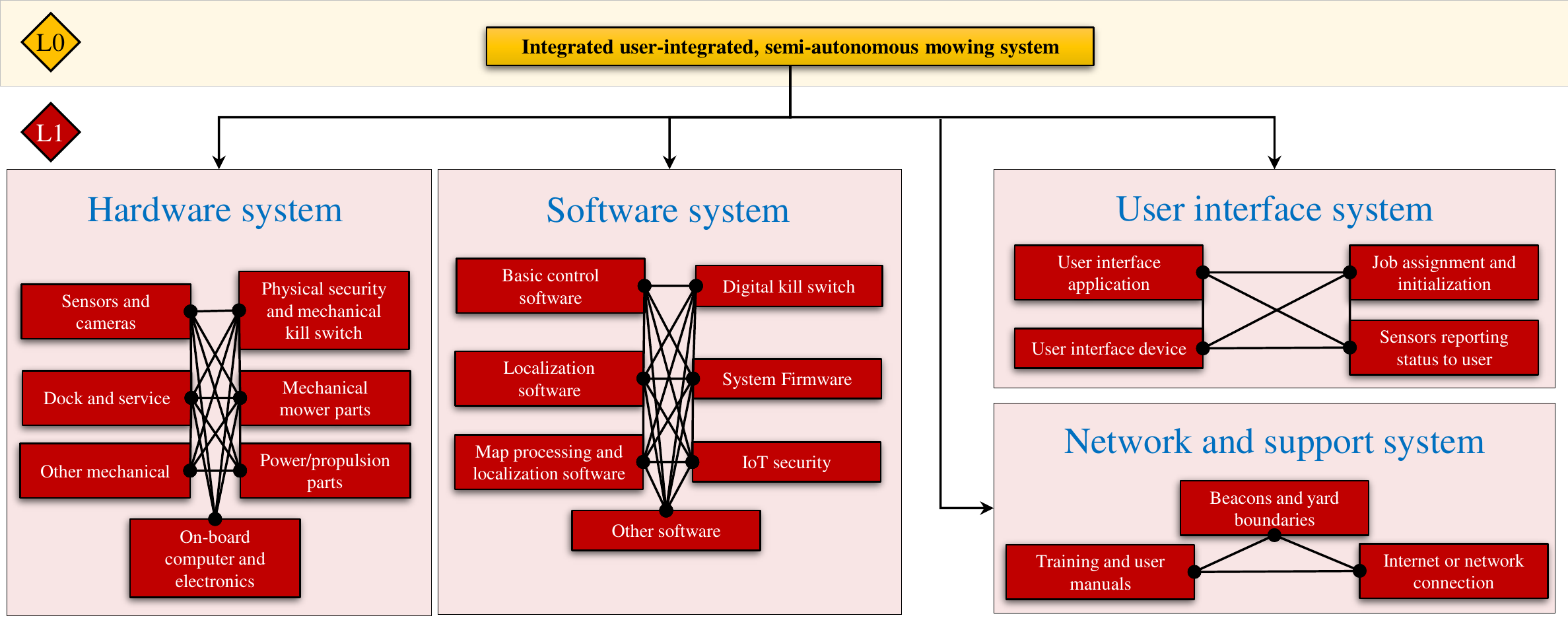}
\caption{Proposed integrated system architecture}
\label{fig:Integrated}
\end{figure}

The integrated system contains the same basic elements and functions as the modular system, but is fully integrated within each of the main system divisions. There are only two levels in this case, since the system is not modular. Note that there may be modular components within each of the main system functions but ideally the system is entirely integrated. Figure~\ref{fig:Integrated} demonstrates an example architecture with each of the major subsystem elements integrated together into the four main system divisions previously discussed.    

\begin{figure}[H]
\centering
\includegraphics[width=1\textwidth]{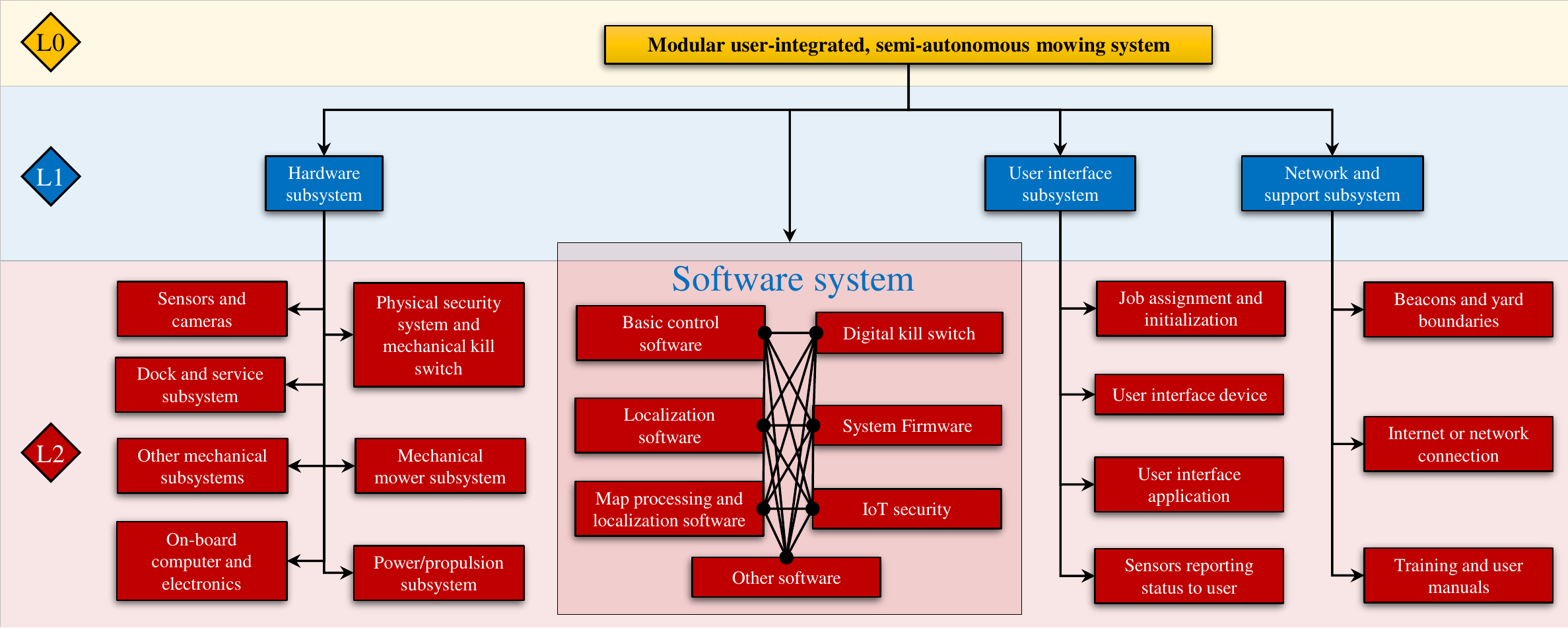}
\caption{Example hybrid system architecture (first three levels) with integrated software system and a modular architecture for the rest of the system}
\label{fig:Hybrid}
\end{figure}

Likely the most useful and practical case would be the hybrid system architecture, which combines elements of both the modular and hybrid cases to find a good balance between system flexibility and optimizability. For example, (Figure~\ref{fig:Hybrid}), it could be decided by the stakeholders to use an integrated software system while making the rest of the system modular. There is a nearly infinite number of combinations possible for a hybrid system, ranging from 99.9\% modular to 99.9\% integrated, so this configuration is shown only for demonstration purposes.  

\section{Conclusions and Final Remarks}
\label{Sec 5}
\vspace{-5pt}
This concept paper laid out some recent work by the authors toward the development of a systems engineering perspective on the design of user-integrated, semi-autonomous mowing systems. The paper began with an extensive literature review to find the current status of development for this kind of system, with the conclusion that the literature is sparse and does not seem to be focused around any kind of standard method or perspective. To address this, a standardized system-level perspective was proposed, both for requirements definition and for the system architecture. For the system architecture, three different arrangements were examined and discussed relative to each other, namely the modular architecture, the integrated architecture, and the hybrid of the two. These three all have various advantages and disadvantages and the best choice depends on the system characteristics and detailed requirements, but a proposed arrangement (with interfaces marked) of the essential system elements was given for each.

This work is meant to provoke a discussion and interest in the systems engineering and robotics communities and help to start developing a standardized and consistent perspective for the future development of refinement of these and other kinds of user-integrated semi-autonomous systems. This work should serve as a starting place for the research community (both commercial and academic) to improve and develop a systems engineering method for these systems, hopefully culminating in a universal design approach and useful industry standards related to semi-autonomous and autonomous systems in the future.

\mbox{}
 
\nomenclature{CONOPS}{~~~Concept of operations} 
\nomenclature{GPS}{~~~~~~~~Global positioning system}
\nomenclature{IoT}{~~~~~~~~Internet of things}
\nomenclature{LIDAR}{~~~~~~~Light detection and ranging}
\nomenclature{PNIO}{~~~~~~~~Predictable non-interfering obstacle}
\nomenclature{PNO}{~~~~~~~~Predicable interfering obstacle}
\nomenclature{UAV}{~~~~~~~~Unmanned aerial vehicle}
 
\printnomenclature

\end{document}